\pdfoutput=1

\documentclass[11pt]{article}

\usepackage{acl}

\usepackage{times}
\usepackage{latexsym}
\usepackage{longtable}

\usepackage{titling}
\usepackage{titlesec}
\usepackage{etoolbox}
\usepackage[T1]{fontenc}

\usepackage[utf8]{inputenc}

\usepackage{microtype}

\usepackage[utf8]{inputenc} 
\usepackage[T1]{fontenc}    
\usepackage{hyperref}       
\usepackage{url}            
\usepackage{booktabs}       
\usepackage{amsfonts}       
\usepackage{nicefrac}       
\usepackage{microtype}      

 \usepackage{graphicx}
\usepackage{subfigure} 
\usepackage{pifont}
\usepackage{adjustbox}
\usepackage{lipsum}
\usepackage{color}
\usepackage{wrapfig}
\usepackage{booktabs}
\usepackage[textsize=scriptsize]{todonotes}
\usepackage{multirow,mathtools}
\usepackage{threeparttable}
\usepackage{pifont}

\definecolor{lightblue}{rgb}{0.68, 0.85, 0.9}
\definecolor{lightgreen}{rgb}{0.56, 0.93, 0.56}
\definecolor{lightskyblue}{rgb}{0.53, 0.81, 0.98}
\definecolor{non-photoblue}{rgb}{0.64, 0.87, 0.93}
\definecolor{magicmint}{rgb}{0.67, 0.94, 0.82}
\definecolor{mossgreen}{rgb}{0.68, 0.87, 0.68}
\definecolor{salmon}{rgb}{1.0, 0.55, 0.41}
\definecolor{babypink}{rgb}{0.96, 0.76, 0.76}
\definecolor{darkgreen}{rgb}{0, 0.7, 0}

\DeclareMathAlphabet\mathbfcal{OMS}{cmsy}{b}{n}

\usepackage{color, colortbl}
\definecolor{Gray}{gray}{0.93}
\definecolor{Orange}{rgb}{1,0.5,0}
\definecolor{DGray}{gray}{0.83}
\definecolor{LightCyan}{rgb}{0.88,1,1}

\usepackage[most]{tcolorbox}
\newtcolorbox{mybox}[2][]{%
  attach boxed title to top center
               = {yshift=-8pt},
  colback      = Gray,
  colframe     = black,
  fonttitle    = \bfseries,
  colbacktitle = white,
  title        = #2,#1,
  enhanced,
}

\DeclarePairedDelimiterX{\inp}[2]{\langle}{\rangle}{#1, #2}

\makeatletter
\newcommand*{\rom}[1]{\expandafter\@slowromancap\romannumeral #1@}
\makeatother
\newcommand{\mycomment}[1]{}
\title{Leveraging LLMs for Dialogue Quality Measurement}

\author{Jinghan Jia$^{1}$, Abi Komma$^{2}$, Timothy Leffel$^{2}$,  Xujun Peng$^{2}$\\
        \textbf{Ajay Nagesh$^{2}$, Tamer Soliman$^{2}$, Aram Galstyan$^{2}$, Anoop Kumar$^{2}$} \\
        \textsuperscript{\normalfont 1} Computer Science \& Engineering, Michigan State University \\
        \textsuperscript{\normalfont 2} Amazon AGI Foundations \\
            \texttt{   jiajingh@msu.edu, \{kommaak, leffelt, penxujun,  nagesajg,} \\  \texttt{ tsoliman, argalsty\}@amazon.com, anoopkum@gmail.com}
  }

\begin{document}
\maketitle

\begin{abstract}

In task-oriented conversational AI evaluation, unsupervised methods poorly correlate with human judgments, and supervised approaches lack generalization. Recent advances in large language models (LLMs) show robust zero-shot and few-shot capabilities across NLP tasks. This paper explores using LLMs for automated dialogue quality evaluation, experimenting with various configurations on public and proprietary datasets. Manipulating factors such as model size, in-context examples, and selection techniques, we examine ``chain-of-thought'' (CoT) reasoning and label extraction procedures. Our results show that (1) larger models yield more accurate dialogue labels; (2) algorithmic selection of in-context examples outperforms random selection; (3) CoT reasoning where an LLM is asked to provide justifications before outputting final labels improves performance; and (4) fine-tuned LLMs outperform out-of-the-box ones. Our results indicate that LLMs that are suitably fine--tuned and have sufficient reasoning capabilities can be leveraged for automated dialogue evaluation.

\end{abstract}
\section{Introduction}
Evaluating conversational system performance in NLP is challenging. Automating effective evaluation is crucial for enhancing dialogue systems. However, automatic metrics like BLEU \citep{papineni2002bleu} and ROUGE  \citep{lin2004rouge} fall short in accurately measuring perceived quality due to complex mappings \citep{liu2016not}. Advanced methods (USR  \citep{mehri-eskenazi-2020-usr}, FED \citep{mehri2020unsupervised}, DialogRPT \citep{gao2020dialogue}) address this but often require extensive training data and human references, making them costly and limited in generalization to new datasets.

Recent advancements in Large Language Models (LLMs) \citep{bubeck2023sparks} have demonstrated robust zero- or few-shot capabilities and reasoning skills across a range of tasks \citep{brown2020language}. Consequently, researchers have begun to explore the application of LLMs to classification problems such as dialogue evaluation \citep{lin2023llm,huynh2023understanding}. 

Recent studies \citep{lin2023llm,huynh2023understanding} demonstrate that LLMs excel on diverse datasets. However, uncertainties persist regarding the impact of factors like model size and in-context examples on open-sourced LLM performance. This paper aims to clarify these influences. Additionally, there is a lack of comprehensive literature on deploying LLMs for dialogue evaluation. To fill this gap, we propose two common evaluation strategies, providing a comparative analysis of their pros and cons.


In this paper, we systematically study different aspects of LLM-based dialogue evaluation by conducting extensive experiments on two benchmark datasets, one publicly available and the other proprietary, Amazon-internal datasets. We initially explore the connection between different attributes such as model size and in-context examples, and their impact on dialogue evaluation performance. Additionally, we present a dialogue evaluation that leverages  ``chain-of-thought'' (CoT) reasoning abilities of LLMs~\citep{CoT, wang-etal-2023-towards}. 

Our experiments demonstrate that larger model sizes and instruction tuning generally helps with zero shot dialogue evaluation. Furthermore, in few-shot scenario, we find that algorithmic selection of in-context examples yields better results than random selection. Next, we demonstrate that supervised fine-tuning can substantially improve the performance of LLMs on dialogue evaluation task. Finally, we explore and validate a CoT-based evaluation framework which is  capable of returning not only dialogue labels but comprehensive explanations and justifications, thereby offering a more coherent and holistic evaluation. Remarkably, our results indicate that CoT-based evaluation is more accurate when the LLM is prompted to first analyze the dialogue and then produce labels. Combined, our findings confirm the applicability and effectiveness of LLM-based automated dialogue evaluation.

\section{Related work}

\noindent \textbf{Dialogue evaluation.}
Evaluating dialog systems poses challenges like accounting for multiple interlocutors, contextual dynamics, and the one-to-many relationship, as highlighted by \citet{zhang2021dialoguebert} and \citet{zhao2017learning}. Metrics such as USR and FED address these challenges, showing strong correlation with human evaluation standards. Utilizing models like RoBERTa and DialoGPT, notable for their smaller yet effective versions fine-tuned for specific dialog tasks, these metrics excel in capturing nuanced dialog attributes. Other evaluation metrics such as GRADE and DEB~\citep{huang2020grade,mehri2020unsupervised} attempt to measure text coherence, response diversity, engagement, and common sense. With the exponential growth in parameter count of contemporary LLMs and their promising generalization capabilities in NLP tasks, it is anticipated that these model-centric evaluative metrics will undergo further enhancements.

\noindent \textbf{Large language models for evaluations.}
Recent studies explore LLMs in dialogue evaluation. GPTScore \citep{fu2023gptscore} uses models such as GPT-3 \citep{brown2020language}, assigning higher probabilities to superior-quality content and employing diverse prompts for holistic evaluation. Similarly, \citet{huynh2023understanding} investigate using ChatGPT and InstructGPT \citep{ouyang2022training} for reference-independent text quality assessment, contrasting various LLM methodologies, including explicit scoring, leveraging model confidence for implicit score allocation, and direct pairwise text comparison.

The G-EVAL framework \citep{liu2023gpteval} is a notable advancement, synergistically integrating LLMs with the chain-of-thought (CoT) paradigm and a form-filling strategy. Notably, using GPT-4 \citep{bubeck2023sparks} as its foundational model, G-EVAL shows strong correlation with human evaluations in summarization tasks.

\noindent \textbf{Parameter-efficient fine-tuning (PEFT).}
As base language models grow in size \cite{touvron2023llama,zhang2022opt}, researchers frequently turn to parameter-efficient fine-tuning techniques to tailor models for specific downstream tasks. These fine-tuning approaches typically fall into three main categories:

\textbf{(1) Prefix-Tuning:} This method inserts special tokens among input tokens with trainable embeddings for the task at hand \cite{li2021prefix}.

\textbf{(2) Adapter Tuning:} This approach inserts adapter layers between self-attention and MLP modules, providing nuanced control over the model's behavior without altering the core architecture \cite{houlsby2019parameter,zhang2023llama}.

\textbf{(3) Low-Rank Adaptation:} This technique uses trainable low-rank decomposition matrices in each network layer, simplifying the model for efficient fine-tuning \cite{hu2021lora}. It shows promise in adapting large generative models for specific applications \citep{cuenca2023using,zhang2023llama}.

These strategies reflect ongoing efforts to make large-scale models more adaptable and efficient, leveraging their vast capacities while mitigating computational and practical challenges.
\\
\noindent \textbf{In-context learning.}
In-context learning is a prompting technique for LLMs in which example input-output pairs for some task are injected into the prompt before the target input is presented. The idea is that seeing correct examples of the task will help the model to provide a correct target output.

Selecting in-context examples is crucial for effectively prompting LLMs, enabling them to pivot to new tasks without extensive fine-tuning. Examples play a pivotal role in guiding LLMs' predictive capabilities, with research exploring methods such as semantic proximity evaluation \citep{liu2021makes} and retrieval mechanisms such as BM25 \citep{robertson2009probabilistic}, used independently or in an initial training phase for a selector retriever.

These selection approaches excel in few-shot NLP tasks. For instance, in \citet{su2022selective}, a bifurcated framework effectively annotates and selects in-context samples from unlabeled repositories, achieving impressive performance across various tasks. Similarly, \citet{liu2021makes} suggested that choosing examples with congruent sentence embeddings optimizes GPT-3's efficacy. Despite positive outcomes, there is a need for deeper explorations to discover more general in-context example retrieval methodologies.
\section{Methodology}

\begin{figure}[t]
\centerline{
\includegraphics[width=0.9\linewidth]{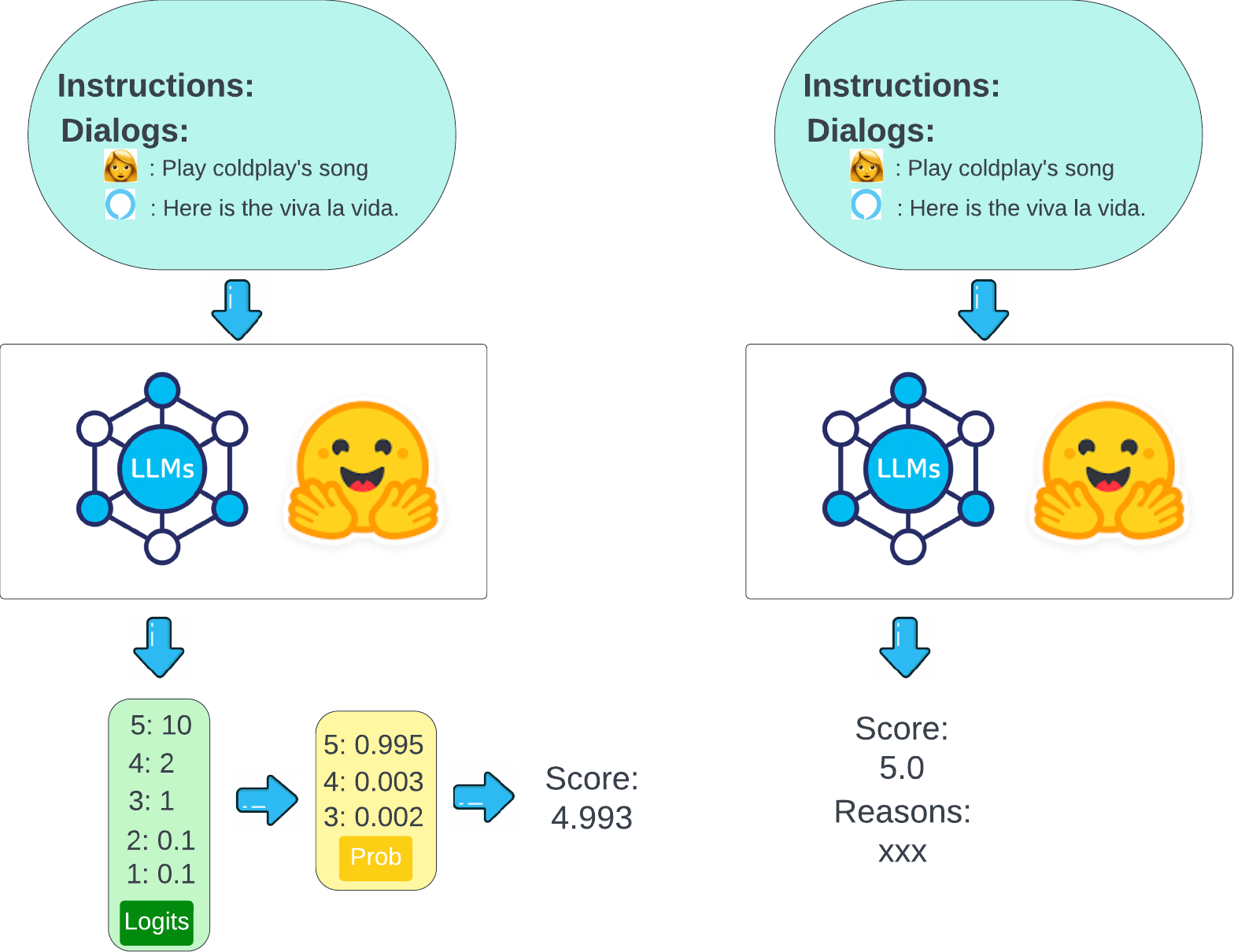}
}
\caption{{
Schematic overview of LLM dialogue evaluation methods. Left: Pipeline using logits method for generating scores from LLMs. Right: Pipeline employing generation method to produce ratings from LLMs.
}}
\label{fig: methodology}
\end{figure}
\noindent \textbf{Dialogue evaluation with logits.}  LLMs like GPT \citep{radford2018improving} with a decoder-only architecture are autoregressive. They generate sequences one element at a time, each conditioned on the preceding ones. The probability of a token sequence $x = (x_1, x_2, ..., x_T)$ is modeled as the product of conditional probabilities for each token given its history.
\begin{align}
    p(x) = \prod_{t=1}^{T} p(x_t | x_{<t})
    \label{eq: llms}
\end{align}
where $x_{<t} = (x_1, ..., x_{t-1})$ is the history before $x_t$, and $p(x_t | x_{<t})$ is typically modeled by a softmax over the vocabulary. LLMs can be prompted to provide a score (example prompt in section \ref{subsec:prompts1}), using the returned probabilities to generate ratings.

Building on methods from \citet{huynh2023understanding}, we use these properties to select the top-$K$ ratings $r_1, r_2,..., r_K$, based on their corresponding log probabilities $p_1, ..., p_K$. We then perform a weighted sum of these ratings, as illustrated in the left panel of Figure~\ref{fig: methodology}. The weights can be calculated using Equation \eqref{eq: weight}:
\begin{align}
    w_i = \frac{p_i}{\sum_{j=1}^K p_j}
    \label{eq: weight}
\end{align}
The final rating can be represented by Equation \eqref{eq: rating}:
\begin{align}
    r = \sum_{i=1}^K r_i\ast w_i
    \label{eq: rating}
\end{align}

\noindent \textbf{Dialogue evaluation with generation.}
In addition to the mentioned method, \citet{lin2023llm} proposed a novel framework where LLMs are prompted to directly generate responses for dialogue evaluation. Ratings for the dialogue can then be extracted from the produced LLM responses. See prompts in section \ref{sec: prompts}.

\section{Experiment setup}

\begin{figure}[h]
\vspace*{-3mm}
\centerline{
\begin{tabular}{ccc}
    \hspace*{-2mm}  \includegraphics[width=25mm,height=!]{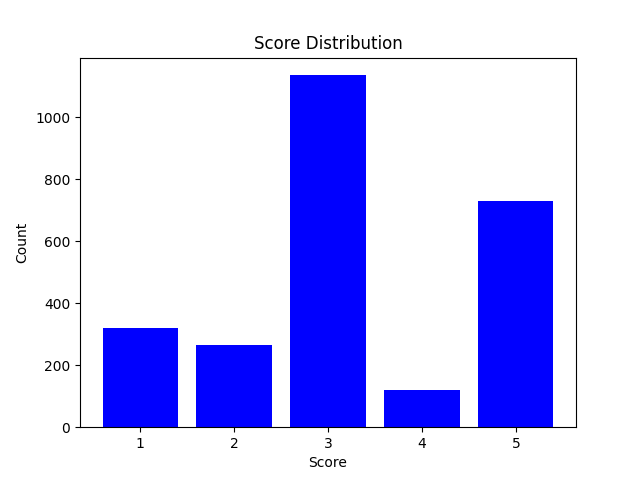} &
    \hspace*{-5mm} \includegraphics[width=25mm,height=!]{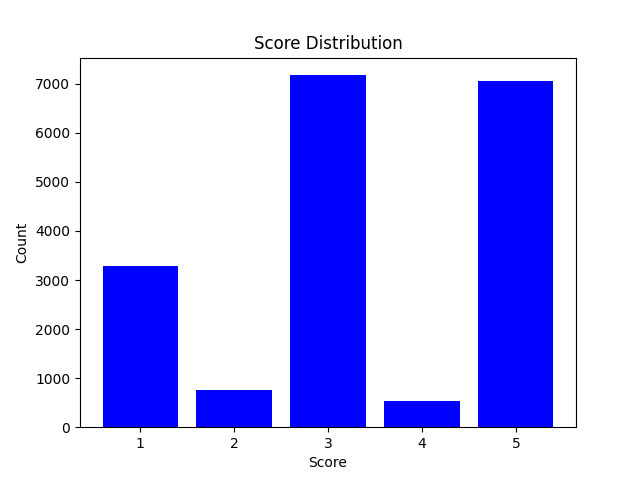} &
    \hspace*{-5mm} \includegraphics[width=25mm,height=!]{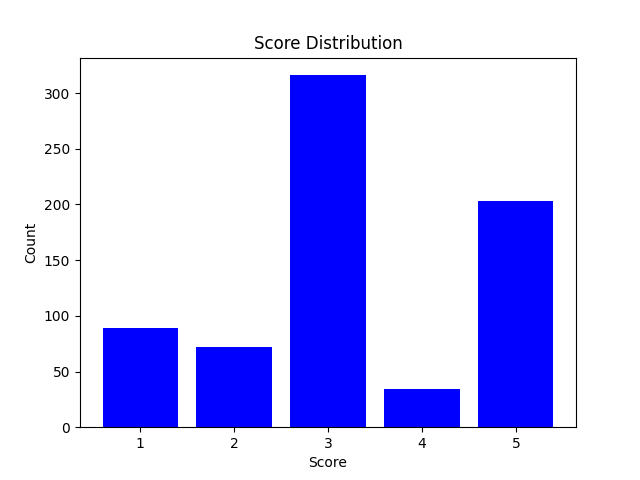} \\
    \hspace*{-2mm} (a) small train & \hspace*{-4mm}(b) large train & \hspace*{-5mm}(c) test
\end{tabular}
}
\vspace*{-2mm}
\caption{ Score distribution in train and test splits from the Amazon-internal dataset.
}
 \label{fig: score_distribution}
 \vspace*{-4mm}
\end{figure}

\noindent \textbf{Model.} 
In this study, we utilize models from the Llama family \citep{touvron2023llama} and the Falcon series~\citep{falcon40b}. We also incorporate the instruction-tuning variants of these models as proposed in the Alpaca study \cite{alpaca}. Temperature is fixed at $0.7$ during generation. \\
\noindent \textbf{Dataset.}
We experiment on two datasets: the publicly available USS \citep{Sun:2021:SUS} dataset includes SGD, MultiWOZ, ReDial, and CCPE subsets, with a 1-5 quality score scale. We randomly allocate $10\%$ as test data, using the rest for training (supervised fine-tuning or in-context learning). We also evaluate our methods on two versions of an Amazon-internal dialogue-quality dataset, which has a human-annotated quality rating on a scale of 1-5 (similar in format to the data described in \citet{komma2023toward}). The rating distribution is shown in Figure \ref{fig: score_distribution}, with the smaller training set's rating distribution resembling the test set more closely than the larger one.

We binarize the datasets, considering scores of three and below as ``defect'' (unsatisfactory) and scores of four or five as ``non-defect'' (satisfactory). This simplified scheme enables us to use standard binary classification metrics to compare the effectiveness of different methods.
\\

\noindent \textbf{In-context example selection methods.}
Here we use three methods to select in-context examples. The first is to simply source examples randomly from the training dataset. The second employs a probabilistic algorithm for information retrieval (IR) \citep{robertson2009probabilistic}, selecting similar examples based on a defined similarity metric. The third relies on BERT \citep{devlin2018bert}, extracting representations from the input dialogue and identifying similar examples using cosine similarity computations. Together, these three methods offer diverse strategies for in-context example selection. We conducted three runs for the random selection method and report the mean value. \\ 

\noindent \textbf{Supervised fine-tuning method.} In this study, we also explore supervised fine-tuning to adapt LLMs to the task of dialogue quality evaluation. To manage cost, we adopt the LoRA setting \citep{hu2021lora}, fine-tuning a relatively small number of parameters compared to full-rank fine tuning. LoRA fine-tuning efficiently enhances the model's effectiveness at the target task while addressing computational challenges.

\section{Results}
We now describe the results of our experiments designed to analyze various aspects of LLM-based dialogue evaluation.
\subsection{Larger models help zero-shot dialogue evaluation.} 
Table\,\ref{tab:model_size} shows the relationship between model size and zero-shot ability in dialogue quality evaluation, comparing Spearman and Pearson correlation values with human annotation for different model configurations across the Alpaca, Llama, and Falcon series.

Table\,\ref{tab:model_size} suggests a positive relationship between model size and zero-shot ability in dialogue quality evaluation. Notably, the Falcon series shows a significant improvement (Spearman: -0.02 to 0.41 for Falcon-7b versus Falcon-40b). Alpaca 13b sees marginal improvements compard to 7b, but the Llama series exhibits no significant improvement as size increases, with weak correlations across sizes (possibly due to the original Llama's poor instruction understanding).

This conclusion underscores the potential benefits of employing larger models in dialogue systems, particularly for applications that require zero-shot flexibility. 
\begin{table}[t]
    \centering
    \large
    \caption{Performance comparison across model with different sizes on Amazon-internal datasets. Spearman and Pearson correlation values presented, with the best results highlighted in bold.}
    \resizebox{0.49\textwidth}{!}{\begin{tabular}{l|cc|ccc|cc}
    \toprule
    Models                & Alpaca-7b & Alpaca-13b & Llama-7b & Llama-13b & Llama-30b & Falcon-7b-instruct & Falcon-40b-instruct \\
    \midrule
    Spearman correlation   & 0.47      & \textbf{0.48}       & 0.00     & 0.01     & \textbf{0.01}     & -0.02     & \textbf{0.41}       \\
    Pearson correlation    & 0.47      & \textbf{0.48}       & 0.03     & 0.02    & \textbf{0.03}     & -0.02     & \textbf{0.35}       \\
    \bottomrule
    \end{tabular}
    }
    \label{tab:model_size}
    \vspace*{-3mm}
\end{table}

\subsection{Instruction-tuning helps zero-shot dialogue evaluation.}
Table~\ref{tab:model_size} also provides insight into the impact of instruction fine-tuning on models' ability to do zero-shot dialogue quality evaluation. The Alpaca series underwent instruction fine-tuning based on the foundational Llama models, an important modification to enhance task-specific performance. Our empirical results illustrate the superiority of the Alpaca models in the realm of zero-shot dialogue quality evaluation, as evidenced by consistently higher Spearman and Pearson correlation coefficients. This observation underscores the significance of instruction tuning as a critical technological approach for augmenting dialogue evaluation performance. For example, the improvement from Alpaca-13b compared to the Llama-13b models is 0.47 on the Spearman correlation shown in Table~\ref{tab:model_size}~\footnote{Llama-7b having a Spearman correlation value of 0 is actually rounded value of a tiny correlation of < 0.005.}. We hypothesize that instruction tuning may serve to refine LLMs comprehension of instructions or prompts, thereby optimizing their ability to execute the specified tasks with greater accuracy. The results suggest that such tuning may be instrumental in facilitating a more nuanced understanding of dialogues, opening avenues for further research and development in this domain.

\begin{table*}[t]
\centering
\caption{Comparison of different in-context example selection methods and zero-shot ability for dialogue quality evaluation across different datasets using Falcon-7b-instruct Model on open-source dataset.}
\resizebox{0.95\textwidth}{!}{
\begin{tabular}{lcccccccccccccccc}
\toprule
Dataset & Model & Defect Rate & \multicolumn{3}{c}{Defect Class} & \multicolumn{3}{c}{Non-Defect Class} & \multicolumn{3}{c}{Weighted Average} & \multicolumn{3}{c}{Macro Average} & {Spearman} & Pearson \\
        &       &             & Precision & Recall & F1-Score & Precision & Recall & F1-score & Precision & Recall & F1-score & Precision & Recall & F1-score\\
\midrule
\multirow{7}{*}{CCPE} & Zero-shot & 62\% & 0.71 & 0.39 & 0.5 & 0.42 & 0.74 & 0.54 & 0.6 & 0.52 & 0.51 & 0.57 & 0.56 & 0.52 & 0.21 & 0.27 \\
&Random-1 & 62\% & 0.75 & 0.19 & 0.31 & 0.4 & \textbf{0.89} & 0.56 & 0.62 & 0.46 & 0.4 & 0.58 & 0.54 & 0.43 & \textbf{0.44} & 0.25 \\
&BM25-1 & 62\% & \textbf{0.79} & 0.35 & 0.49 & 0.44 & 0.84 & \textbf{0.58} & 0.66 & 0.54 & 0.52 & 0.62 & 0.6 & 0.54 & 0.4 & \textbf{0.31} \\
&BERT-1 & 62\% & 0.73 & 0.26 & 0.38 & 0.41 & 0.84 & 0.55 & 0.61 & 0.48 & 0.45 & 0.57 & 0.55 & 0.47 & 0.18 & 0.19 \\
&Random-4 & 62\% & 0.71 & 0.55 & 0.62 & 0.46 & 0.63 & 0.53 & 0.61 & 0.58 & 0.59 & 0.58 & 0.59 & 0.58 & 0.16 & 0.17 \\
&BM25-4 & 62\% & 0.73 & \textbf{0.77} & \textbf{0.75} & \textbf{0.59} & 0.53 & 0.56 & \textbf{0.67} & \textbf{0.68} & \textbf{0.68} & \textbf{0.66} & \textbf{0.65} & \textbf{0.65} & 0.23 & 0.2 \\
&BERT-4 & 62\% & 0.68 & 0.68 & 0.68 & 0.47 & 0.47 & 0.47 & 0.6 & 0.6 & 0.6 & 0.58 & 0.58 & 0.58 & -0.07 & -0.13 \\
\midrule
\multirow{7}{*}{MWOZ} &Zero-shot & 55\% & 0.5 & 0.29 & 0.37 & 0.43 & 0.64 & 0.51 & 0.47 & 0.45 & 0.43 & 0.46 & 0.47 & 0.44 & -0.05 & 0.0 \\
&Random-1 & 55\% & 0.55 & 0.11 & 0.18 & 0.45 & \textbf{0.89} & \textbf{0.6} & 0.5 & 0.46 & 0.37 & 0.5 & 0.5 & 0.39 & 0.09 & \textbf{0.1} \\
&BM25-1 & 55\% & 0.53 & 0.15 & 0.23 & 0.45 & 0.84 & 0.58 & 0.49 & 0.46 & 0.39 & 0.49 & 0.49 & 0.41 & \textbf{0.18} & 0.06 \\
&BERT-1 & 55\% & \textbf{0.63} & 0.18 & 0.28 & 0.46 & 0.87 & \textbf{0.6} & 0.55 & 0.49 & 0.43 & 0.54 & 0.52 & 0.44 & 0.13 & 0.0 \\
&Random-4 & 55\% & 0.53 & 0.35 & 0.42 & 0.44 & 0.62 & 0.51 & 0.49 & 0.47 & 0.46 & 0.48 & 0.48 & 0.47 & 0.09 & 0.08 \\
&BM25-4 & 55\% & 0.61 & \textbf{0.6} & \textbf{0.61} & \textbf{0.52} & 0.53 & 0.53 & \textbf{0.57} & \textbf{0.57} & \textbf{0.57} & \textbf{0.57} & \textbf{0.57} & \textbf{0.57} & 0.08 & 0.04 \\
&BERT-4 & 55\% & 0.56 & 0.58 & 0.57 & 0.47 & 0.44 & 0.45 & 0.52 & 0.52 & 0.52 & 0.51 & 0.51 & 0.51 & 0.06 & 0.05 \\
\midrule
\multirow{7}{*}{Redial} & Zero-shot & 44\% & 0.49 & \textbf{0.55} & \textbf{0.52} & \textbf{0.61} & 0.55 & 0.58 & 0.56 & 0.55 & 0.55 & 0.55 & 0.55 & \textbf{0.55} & 0.09 & 0.11 \\
&Random-1 & 44\% & 0.5 & 0.2 & 0.29 & 0.57 & \textbf{0.84} & 0.68 & 0.54 & 0.56 & 0.51 & 0.54 & 0.52 & 0.49 & 0.18 & 0.24 \\
&BM25-1 & 44\% & 0.48 & 0.27 & 0.35 & 0.57 & 0.77 & 0.66 & 0.53 & 0.55 & 0.52 & 0.53 & 0.52 & 0.5 & 0.2 & 0.17 \\
&BERT-1 & 44\% & \textbf{0.58} & 0.32 & 0.41 & \textbf{0.61} & 0.82 & \textbf{0.7} & \textbf{0.6} & \textbf{0.6} & \textbf{0.57} & \textbf{0.59} & \textbf{0.57} & \textbf{0.55} & \textbf{0.24} & \textbf{0.29} \\
&Random-4 & 44\% & 0.5 & 0.41 & 0.45 & 0.59 & 0.68 & 0.63 & 0.55 & 0.56 & 0.55 & 0.55 & 0.54 & 0.54 & 0.12 & 0.11 \\
&BM25-4 & 44\% & 0.4 & 0.45 & 0.43 & 0.52 & 0.46 & 0.49 & 0.47 & 0.46 & 0.46 & 0.46 & 0.46 & 0.46 & 0.1 & 0.15 \\
&BERT-4 & 44\% & 0.44 & \textbf{0.55} & 0.49 & 0.57 & 0.46 & 0.51 & 0.51 & 0.5 & 0.5 & 0.5 & 0.5 & 0.5 & 0.05 & 0.05 \\
                       \midrule
\multirow{7}{*}{SGD} & Zero-shot & {48\%} & \textbf{0.64} & \textbf{0.63} & \textbf{0.63} & \textbf{0.66} & {0.67} & \textbf{0.67} & \textbf{0.65} & \textbf{0.65} & \textbf{0.65} & \textbf{0.65} & \textbf{0.65} & \textbf{0.65} & \textbf{0.34} & \textbf{0.32} \\
                      & Random-1  & 48\% & 0.40 & 0.08 & 0.14 & 0.51 & \textbf{0.88} & 0.65 & 0.46 & 0.50 & 0.40 & 0.46 & 0.48 & 0.39 & 0.20 & 0.09 \\
                      & BM25-1    & 48\% & 0.39 & 0.15 & 0.21 & 0.50 & 0.79 & 0.61 & 0.45 & 0.48 & 0.42 & 0.44 & 0.47 & 0.41 & 0.29 & 0.15 \\
                      & BERT-1    & 48\% & 0.50 & 0.15 & 0.23 & 0.52 & 0.87 & 0.65 & 0.51 & 0.52 & 0.45 & 0.51 & 0.51 & 0.44 & 0.33 & 0.26 \\
                      & Random-4  & 48\% & 0.44 & 0.33 & 0.38 & 0.50 & 0.62 & 0.55 & 0.47 & 0.48 & 0.47 & 0.47 & 0.47 & 0.47 & -0.01 & -0.02 \\
                      & BM25-4    & 48\% & 0.52 & 0.58 & 0.55 & 0.57 & 0.50 & 0.53 & 0.54 & 0.54 & 0.54 & 0.54 & 0.54 & 0.54 & 0.06 & 0.08 \\
                      & BERT-4    & 48\% & 0.37 & 0.40 & 0.38 & 0.41 & 0.38 & 0.40 & 0.39 & 0.39 & 0.39 & 0.39 & 0.39 & 0.39 & -0.12 & -0.12 \\

\bottomrule
\end{tabular}
}
\vspace*{-3mm}
\label{tab:in-context-examples}
\end{table*}

\subsection{In-context examples enhance the performance of dialogue evaluation.}
In this section, we evaluate the influence of in-context examples on the base model Falcon-7b-instruct, utilizing three distinct in-context example selection methods: Random, BM25, and BERT. For the BM25 and BERT-based ICL approaches, we selected either 1 or 4 of the most semantically similar samples from the training set. The selection was based on the semantic similarity between the examples, as determined by the BM25 and BERT models. In contrast, for the random-selection ICL approach, we randomly picked either 1 or 4 examples from the training set to use as the in-context examples. These experiments are conducted on the open-sourced dataset USS shown in Table\,\ref{tab:in-context-examples}.

First, we observe that in general, the few-shot performance is better than the zero-shot performance. A closer examination reveals that the best performance across various evaluation metrics is primarily concentrated within the few-shot settings in the first three datasets. If we take a closer look at the MWOZ dataset, all highlighted numbers are in the few-shots settings. This observation substantiates the notion that providing in-context examples can indeed enhance the performance in dialogue evaluation. However, it is worth noting that an excessive provision of examples does not necessarily lead to further improvement. We hypothesize that this limitation may stem from the capacity constraints of LLMs, which can struggle to process overly lengthy inputs, occasionally resulting in performance degradation. This pattern is particularly evident in the Redial dataset, where the performance of the 4-shot approach does not surpass the results obtained from the 1-shot experiment. In addition to these findings, Table~\ref{tab:in-context-examples} reveals that zero-shot outperforms the few-shot settings in the SGD dataset. This is likely attributable to the capacity constraints of LLMs; the dialogue lengths should not be excessively long. Notably, the dialogues in the SGD dataset have more turns ($26.7$ turns per dialogue) compared to the other three datasets from \citet{Sun:2021:SUS}. Such findings further emphasize the nuanced relationship between the number of in-context examples and the resulting performance, highlighting the importance of careful selection in few-shot learning.

Second, when comparing the performance across different in-context selection methods, we find that algorithmic selection methods result in notable performance improvements over random selection for in-context examples. For instance, the BM25 and BERT methods consistently perform best across all datasets. Upon closer examination of the CCPE dataset in Table~\ref{tab:in-context-examples}, we observe that $85\%$ of the highest values across all metrics are derived from the algorithm's selected method. What is more, the optimal choice for selecting different in-context example methods varies based on the dataset. In the first two datasets, the BM25 method excels, while in the third, the BERT method stands out.

In conclusion, our results suggest that in-context examples can significantly enhance the quality of dialogue evaluation. Our findings also underscore the importance of employing algorithmic methods for selecting these examples, as the right selection strategy can lead to meaningful performance gains. By revealing these patterns, our study contributes to a deeper understanding of how few-shot learning can be best utilized in dialogue systems.
\begin{table}[hbt!]
\centering
\caption{Summary of performance metrics for supervised finetuning models on Amazon-internal datasets across different training datasets and various model architectures.}
\vspace*{-3mm}
\resizebox{0.49\textwidth}{!}{
\begin{tabular}{lccccccc}
\hline

Models           & Spearman & Pearson & Precision& Recall & F1\ & F1-micro \\
\hline
Alpaca-7b          & 0.47     & 0.47    & 0.52             & \textbf{0.72}          & 0.60      & \textbf{0.69} \\
Alpaca-7b-sft-small & 0.61     & 0.61    & \textbf{0.96}             & 0.59          & \textbf{0.73}      & {0.65} \\
Alpaca-7b-sft-large & \textbf{0.64}     & \textbf{0.66}    & 0.93             & 0.58          & 0.72      & 0.33 \\
\midrule
Llama-7b           & 0.00     & 0.03    & 0.67             & \textbf{1.00}          & \textbf{0.80}      & 0.33 \\
Llama-7b-sft-small & 0.58     & 0.60    & 0.92             & 0.62          & 0.74      & \textbf{0.58} \\
Llama-7b-sft-large & \textbf{0.64}     & \textbf{0.66}    & 0.93             & 0.58          & 0.72      & 0.33 \\
\midrule
Llama-13b          & 0.01    & 0.02   & 0.67             & \textbf{1.00}          & \textbf{0.80}      & 0.36 \\
Llama-13b-sft-small & 0.64     & 0.65    & 0.94             & 0.61          & 0.74      & \textbf{0.46} \\
Llama-13b-sft-large & \textbf{0.64}     & \textbf{0.65}    & 0.97             & 0.54          & 0.69      & 0.40 \\
\midrule
Falcon-40b-instruct         & 0.41     & 0.36    & \textbf{1.00}             & 0.01          & 0.02      & 0.67 \\
Falcon-40b-instruct-sft-small   & 0.60     & 0.61    & 0.96             & 0.52          & 0.67      & 0.65 \\
Falcon-40b-instruct-sft-large   & \textbf{0.61}     & \textbf{0.63}    & 0.93             & \textbf{0.53}          & \textbf{0.68}      & \textbf{0.69} \\
\midrule
\hline
\end{tabular}}

\label{tab:sft_results}
\end{table}

\begin{table*}[!h]
\centering
\small
\caption{\footnotesize{Comparison of CoT methods on the internal dataset over Falcon-7b-instruct. ``Rating-first'' refers to generating the score first, followed by the reasons, while ``Analysis-first'' involves generating an analysis first, then determining the scores.}}
\vspace*{-3mm}
\resizebox{0.99\textwidth}{!}{
\begin{tabular}{lccccccccccccccc}
\toprule
Model & Defect Rate & \multicolumn{3}{c}{Defect Class} & \multicolumn{3}{c}{Non-Defect Class} & \multicolumn{3}{c}{Weighted Average} & \multicolumn{3}{c}{Macro Average} & Spearman & Pearson \\
 & & Precision & Recall & F1-Score & Precision & Recall & F1-score & Precision & Recall & F1-score & Precision & Recall & F1-score & & \\
\midrule
Rating-first & 59\% & 0.59 & \textbf{1.00} & 0.74 & \textbf{0.67} & 0.01 & 0.02 & 0.62 & 0.59 & 0.45 & 0.63 & 0.50 & 0.38 & 0.10 & 0.07 \\
Analysis-first & 59\% & \textbf{0.67} & {0.86} & \textbf{0.75} & 0.66 & \textbf{0.39} & \textbf{0.49} & \textbf{0.67} & \textbf{0.67} & \textbf{0.65} & \textbf{0.67} & \textbf{0.63} & \textbf{0.62} & \textbf{0.23} & \textbf{0.28} \\
\bottomrule
\end{tabular}}
\vspace*{-3mm}
\label{tab:resaons_ratings}
\end{table*}

\subsection{Supervised fine-tuning improves dialogue evaluation quality}
Here we examine the influence of supervised fine-tuning (SFT) on the performance of LLMs for the dialogue evaluation task. Specifically, we fine-tuned the models using both Likert-scale and binary label data. We leverage two internal datasets used for training dialogue quality estimation models. The first is a small version of the dataset, resulting in models denoted as ``xxx-small,'' while the second is a larger version, leading to models labeled as ``xxx-large.''

We report on several classification metrics, including precision, recall, F1-score, and F1-micro. Our findings from Table\,\ref{tab:sft_results} reveal a consistent improvement in the original model's performance after SFT, especially in the Spearman and Pearson correlations with human annotation. For instance, the model ``Falcon-40b-instruct-sft-large'' exhibits a $48\%$ relative improvement compared to the original ``Falcon-40b-instruct.'' This indicates that SFT enhances the alignment of the model's scoring with human evaluation.

Analysis of F1-micro shows that SFT generally improves performance in comparison to the original model. When comparing the Llama-7b-sft-small to the original Llama-7b, there is a $75\%$ relative improvement. A closer examination also suggests that utilizing a larger dataset can boost overall performance. For example, the highest correlations in the first two columns are consistently associated with models trained on the larger dataset. Interestingly, models trained on smaller datasets occasionally exhibit superior F1-micro scores, as observed for the Llama series. For example, when comparing the Llama-13b-sft-small to the original Llama-13b-sft-large, there is a $15\%$ relative improvement. We hypothesize that this may occur when the score distribution between the small dataset and the test dataset aligns more closely shown in Figure~\ref{fig: score_distribution}.

It is important to note that Llama-7b and Llama-13b predict defects for all test samples, which results in a recall of $1.0$ and a precision of $0.67$---this obviously does not indicate optimal performance. To better understand the relationship between model predictions and the ground truth, we should consider the Spearman and Pearson correlations over likert scores, which provide more insight into the linear relationship between the predictions and the human labels.

In conclusion, our findings show that SFT can substantially enhance the performance of LLMS on dialogue evaluation. This study underscores the value of fine-tuning and dataset selection in achieving more accurate and human-aligned evaluations in the context of dialogue systems.

\subsection{Chain-of-thought for generation of scores and reasons} 

In this section, we shift our focus to the ``generation and chain-of-thoughts'' approach \cite{CoT}, which not only generates scores but also provides natural language {\it reasons} for selecting those scores. We explore two distinct paradigms to accomplish this task. The first paradigm, \textbf{Analysis-first}, involves prompting the model to generate an analysis first and then derive ratings based on that analysis. The second paradigm, \textbf{Rating-first}, prompts the model to generate a rating first and then elucidate the reasons for choosing that particular score (see Section~\ref{sec: prompts} for respective prompts).

Interestingly, our findings suggest that the first paradigm---Analysis-first---provides more aligned scores and reasons, as shown in Table \ref{tab:resaons_ratings}. We observe consistent improvement compared to the Rating-first approach. For example, Analysis-first methods outperform Rating-first in $85\%$ of all evaluation metrics. Upon conducting a failure analysis, we discovered that for Rating-first, the scores do not always align with the subsequent reasons. However, in the Analysis-first paradigm, there is consistent alignment between ratings and scores. We attribute the observed metric improvement to this consistency.

The implications of these findings may extend to various applications where the alignment between scores and reasoning is essential. Further exploration of these paradigms and their potential advantages and limitations may provide valuable insights into the optimal utilization of LLMs for complex tasks such as rating and explanation generation. 

\section{Conclusion}
This paper explores the application of LLMs to evaluation of task-oriented dialogue systems. Key findings include: the impact of pretrained model size; the importance of instruction fine-tuning; the effectiveness of in-context examples; consistent performance improvement through supervised fine-tuning; and that the chain-of-though paradigm is most effective with the Analysis-first approach. 
\section{Limitations}
Our experiments provide valuable insights, but there are limitations. We focus on open-sourced models, excluding closed ones like ChatGPT and Claude. Evaluation primarily centers on user satisfaction, lacking metrics for interestingness and coherence. Performance is influenced by prompt designs, and suboptimal prompts may lead to decline. 

\section{Ethics Statement}
We acknowledge ethical concerns in using LLMs in our evaluation. Firstly, LLMs may carry biases that could impact dialogue evaluation negatively. Secondly, our focus on user satisfaction might overlook issues like toxic responses, leading to inadequate evaluations. Lastly, concerns arise about unintentional release of private information during reason and rating generation. To address these concerns, researchers should exercise caution in using LLMs for reasons and ratings, ensuring accuracy and fairness in interpretation.
\bibliography{anthology,custom}

\newpage
\appendix

\section{Prompts}
In this section, we elaborate on the prompts and instructions we used in the logits-based and generation-based evaluation methods. 
\subsection{Prompts for logits method} \label{subsec:prompts1}
The prompts we used in the logits method are formulated in the following manner:
\vspace*{-3mm}
\begin{tcolorbox}[width=\columnwidth,colback=white]
    \small
    \textbf{Instruction:}
    Could you please evaluate the subsequent dialogue by assigning a score from the given set [1,2,3,4,5]? A score of 1 implies dissatisfaction, while a 5 signifies high satisfaction.
\end{tcolorbox}

\subsection{Prompts for generation method}
\noindent \textbf{Rating-first}
In this section, we outline the prompts utilized in the Rating-First generation method. This process begins with an instruction that directs the Large Language Models (LLMs) to provide both reasons and ratings for a given dialogue. To enhance the LLMs' comprehension, we also supply evaluation criteria standards. Finally, we detail the evaluation steps necessary to complete the entire procedure for dialogue evaluation. The prompt is shown as follows:
\vspace*{-3mm}
\begin{tcolorbox}[width=\columnwidth,colback=white]
    \small{
    \textbf{Instruction}: Could you please evaluate 
    the subsequent dialogue, providing a score 
    from the set [1,2,3,4,5], and give an explanation 
    for choosing that score?\\
        \textbf{To help you better evaluate, here is the evaluation Criteria:}\\
        A score of 1 means very dissatisfied, where the user repeatedly has to stop or cancel bad responses and repeat their request again;\\
        A score of 2 means dissatisfied, where None of the user goals are achieved,the user expresses negative feedback,steps towards a user goal succeeds but the goal fails;\\
        A score of 3 means normal, where At least one of a user goals succeed, and no negative feedback\\
        A score of 4 means satisfied, where the majority of turns succeeded or moved the user closer to their goal\\
        A score of 5 means very satisfied, all turns either succeeded or moved the user closer to their goal(s), and the user\\ expressed no dissatisfaction, and goal was achieved without unnecessary steps\\
        \textbf{Steps to conduct the evaluation are:}\\
        1.Read the dialog, and the response carefully\\
        2.Rate the response on a scale of 1-5 for satisfaction level from user, according to the criteria above\\
        3.Provide a brief explanation for your rating, referring to specific aspects of the response and the dialog.\\}
\end{tcolorbox}

\noindent \textbf{Analysis-first}
In this section, we describe the prompts used in the Analysis-First generation method. Similar to the Rating-First approach, we begin by providing instructions for the LLMs to carry out dialogue evaluation. We then present specific aspects for the LLMs to analyze. To facilitate better understanding, we also supply the evaluation criteria. Finally, we detail the specific steps for the LLMs to follow.
\vspace*{-3mm}
\begin{tcolorbox}[width=\columnwidth,colback=white]
    \small{
    \textbf{Instruction}: Could you please evaluate the subsequent dialogue overall quality, 
    first analysis the dialogue from the following aspects:\\
    1.User goal. \\
    2.User feedback.\\
    3.System response.\\ 
    4.System feedback. \\
    Based the above analysis provide a user satisfactory score from the set [1,2,3,4,5]\\
        \textbf{To help you better evaluate, here is the evaluation Criteria:}\\
        A score of 1 means very dissatisfied, where the user repeatedly has to stop or cancel bad responses and repeat their request again;\\
        A score of 2 means dissatisfied, where None of the user goals are achieved,the user expresses negative feedback,steps towards a user goal succeeds but the goal fails;\\
        A score of 3 means normal, where At least one of a user goals succeed, and no negative feedback\\
        A score of 4 means satisfied, where the majority of turns succeeded or moved the user closer to their goal\\
        A score of 5 means very satisfied, all turns either succeeded or moved the user closer to their goal(s), and the user expressed no dissatisfaction, and goal was achieved without unnecessary steps\\
        \textbf{Steps to conduct the evaluation are:}\\
        1.Read the dialog, and the response carefully\\
        2.Give some brief analysis from the aspects mentioned before\\
        3.Rate the response on a scale of 1-5 for satisfaction level from user, according to the criteria above and the analysis.}
\end{tcolorbox}
\label{sec: prompts}

\end{document}